\documentclass[sigconf]{acmart}

\usepackage[utf8]{inputenc}
\usepackage{graphicx}
\usepackage[utf8]{inputenc}
\usepackage{tabularx}
\usepackage{adjustbox}
\usepackage{comment}
\usepackage{amsmath}
\usepackage{multirow}
\usepackage{graphicx}
\usepackage{color}
\usepackage{colortbl}
\usepackage{tikz}
\usepackage{pifont}
\usepackage{mathtools}
\usepackage{algorithmic}
\usepackage{graphicx}
\usepackage{blkarray, bigstrut}
\usepackage{caption}
\usepackage{adjustbox}
\usepackage{subcaption}
\usepackage{tabularx}
\usepackage{multirow}
\usepackage{comment}
\usepackage{hyperref}
\usepackage{tikz}
\usepackage{float}
\usepackage{makecell}
\AtBeginDocument{%
  }

\copyrightyear{2023} 
\acmYear{2023} 
\setcopyright{rightsretained} 
\acmConference[WWW '23]{Proceedings of the ACM Web Conference 2023}{April 30--May 04, 2023}{Austin, TX, USA}
\acmBooktitle{Proceedings of the ACM Web Conference 2023 (WWW '23), April 30--May 04, 2023, Austin, TX, USA}
\acmDOI{10.1145/3543507.3583358}
\acmISBN{978-1-4503-9416-1/23/04}

\newcommand{\vq}{\boldsymbol{q}}
\newcommand{\vh}{\boldsymbol{h}}
\newcommand{\vr}{\boldsymbol{r}}
\newcommand{\vt}{\boldsymbol{t}}

\begin{document}
%\pagestyle{plain}
%\setlength\abovedisplayskip{2pt}
\title{Link Prediction with  Attention \texorpdfstring{\\}{} Applied on Multiple Knowledge Graph Embedding Models}
\author{Cosimo Gregucci}
\authornote{Both authors contributed equally to this research.}
\orcid{0000-0002-9636-0996}
\affiliation{%
  \institution{University of Stuttgart}
  \city{Stuttgart}
  \country{Germany}
}
\email{cosimo.gregucci@ipvs.uni-stuttgart.de}

\author{Mojtaba Nayyeri}
\orcid{0000-0002-9177-0312}
\authornotemark[1]
\affiliation{
  \institution{University of Stuttgart}
  \city{Stuttgart}
  \country{Germany}}
\email{mojtaba.nayyeri@ipvs.uni-stuttgart.de}

\author{Daniel Hern\'andez}
\orcid{0000-0002-7896-0875}
\affiliation{
  \institution{University of Stuttgart}
  \city{Stuttgart}
  \country{Germany}
}
\email{daniel.hernandez@ipvs.uni-stuttgart.de}

\author{Steffen Staab}
\orcid{0000-0002-0780-4154}
\affiliation{%
 \institution{University of Stuttgart}
  \institution{University of Southampton}
 \city{Stuttgart}
 \country{Germany}}

%\renewcommand{\shortauthors}{Anonymous Authors}


\begin{abstract}
  Predicting missing links between entities in a knowledge graph is a fundamental task to deal with the incompleteness of data on the Web.
  Knowledge graph embeddings map nodes into a vector space to
  predict new links, scoring them according to geometric criteria.
  Relations in the graph may follow patterns that can be learned, e.g., some relations might be symmetric and others might be hierarchical.
  However, the learning capability of different embedding models varies for each pattern and, so far, no single model can learn all patterns equally well.
  In this paper, we combine the query representations from several models in a unified one to incorporate patterns that are independently captured by each model.
  Our combination uses attention to select the most suitable model to answer each query.
  The models are also mapped onto a non-Euclidean manifold, the Poincaré ball, to capture structural patterns, such as hierarchies, besides relational patterns, such as symmetry. 
  We prove that our combination provides a higher expressiveness and inference power than each model on its own. 
  As a result, the combined model can learn relational and structural patterns.
  We conduct extensive experimental analysis 
  with various link prediction benchmarks showing that the combined model outperforms individual models, including state-of-the-art approaches.
\end{abstract}



\maketitle

\section{Introduction}
In the last few years large knowledge graphs (KGs) (e.g. Wikidata \cite{DBLP:journals/cacm/VrandecicK14}, YAGO \cite{suchanek2007yago}, etc) have emerged to represent complex knowledge in the form of multi-relational directed labeled graphs~\cite{hogan2021knowledge}. KGs attracted great attention in industry~\cite{google-knowledge-graph,DBLP:journals/cacm/NoyGJNPT19,ibm-knowledge-graph,microsoft-knowledge-graph,airbnb-knowledge-graph,amazon-knowledge-graph,uber-knowledge-graph} and academia~\cite{DBLP:journals/cacm/VrandecicK14,suchanek2007yago,caligraph,bio2rdf,DBLP:conf/semweb/AuerBKLCI07}, and became the core part of many artificial intelligence systems, e.g., question answering, etc.

Knowledge graphs are typically stored using the W3C standard RDF (Resource Description Framework)~\cite{rdf} which models graphs as sets of triples $(h, r, t)$ where $h$, $r$, and $t$ represent resources that are described on the Web. 
The link prediction community refers to them as \emph{head entity}, \emph{relation}, and \emph{tail entity}, respectively. 
Each triple corresponds to a known fact involving entities $h$ and $t$ and relation $r$. For example, the fact that Berlin is the capital of Germany is modeled as the triple $(\text{Berlin}, \text{capitalOf}, \text{Germany})$. 

A relevant problem for knowledge graphs, called link prediction, is predicting unknown facts (links) 
based on known facts, and knowledge graph embedding (KGE) is a prominent approach for it. 
To predict links,
KGEs map entities $h$ and $t$ and relations $r$ into elements $\vh$, $\vr$, and $\vt$ in a low-dimensional vector space, and score the plausibility of a link $(h,r,t)$ using a \emph{score function} on $\vh$, $\vr$, and $\vt$. 
Most KGE models \cite{bordes2013translating, sun2019rotate, nayyeri20205, trouillon2016complex, zhang2019quaternion} score a link $(h, r, t)$ by splitting it into the \emph{query} $q=(h, r, ?)$ and the corresponding \emph{candidate answer} $t$. 
The query is embedded to an element in the same space as the candidate answers with a transformation function $\vq = g_{\vr}(\vh)$ that depends on the relation $r$ and is applied to $\vh$.
The score of a link is then a measure of the similarity or proximity between $\vq$ and $\vt$.

\begin{figure}[t]
\includegraphics[width=\hsize]{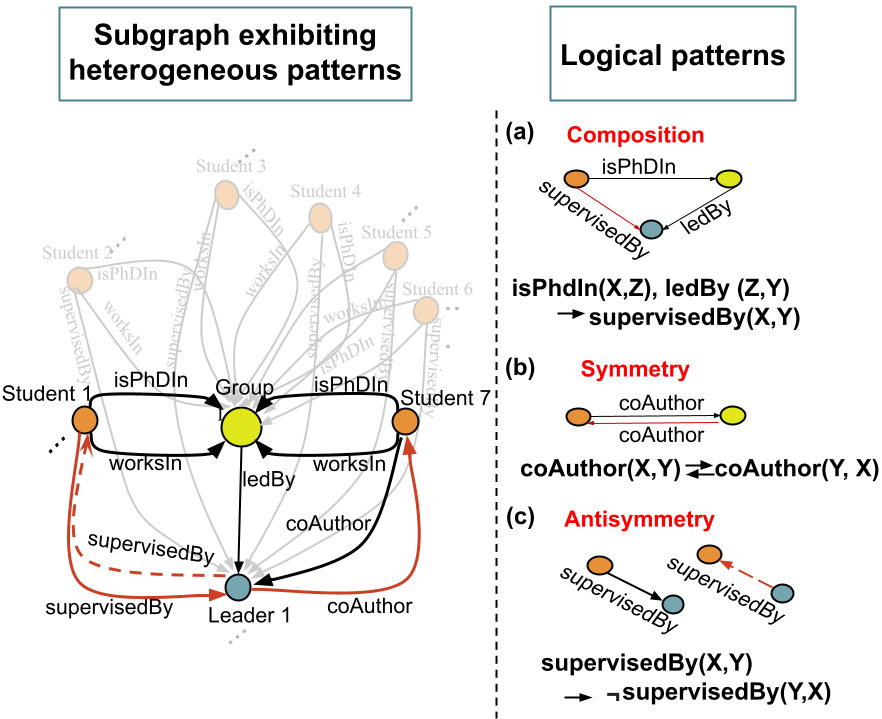}
\centering
\caption{Subgraph exhibiting heterogeneous patterns \cite{nayyeri2021pattern}.}
\label{fig:patterns}
\end{figure}

KGE models can learn logical and other patterns (example in \autoref{fig:patterns}) to predict links. For instance, the facts that co-author is a symmetric relation and part-of is  hierarchical can be learned from the data.
However, the capability of different KGE models to learn and express patterns for predicting missing links varies widely and, so far, no single model does it equally well for each pattern. 
Logical patterns exhibit the form $Premise \xrightarrow{} Conclusion$ where $Premise$ is the conjunction of several atoms and $Conclusion$ is an atom. Structural patterns refer to the arrangements of elements in a graph. A relation forms a hierarchical pattern when its corresponding graph is close to tree-like \cite{chami2020low}, e.g., \textit{(eagle, type-of, bird)}. 
For example, RotatE defines transformations as rotations $g_r^{\mbox{RotatE}}(\vh) = \boldsymbol{h \circ r}$ in Complex space ($\circ$ is an element-wise complex product).
In this way, RotatE can enforce both $\boldsymbol{h \circ r = t}, \boldsymbol{t \circ r = h}$ if $\boldsymbol{r^2 = 1}$
and, thus, it is able to model symmetric relations. 
In Table \ref{tbl:baselines},
we present a summary of the query representation of some state-of-the-art baselines.
We indicate whether a KGE %model
can or cannot model a specific pattern. If it can model a pattern, we further include the number of constraints they have to satisfy to express this pattern. For instance, antisymmetry for RotatE requires two constraints $\boldsymbol{r \neq -1}$ and $\boldsymbol{r \neq 1}$ to be expressed.
 Further explanation of Table \ref{tbl:baselines}
can be found in Appendix \ref{explaintbl1}. 

Beyond the KGEs surveyed in Table 1, further works have defined query representations successfully dealing with different subsets of patterns, such as 5*E \cite{nayyeri20205}, AttE/H \cite{chami2020low}, TransH \cite{wang2014transh}, or ProjE     
\cite{shi2017proje}. However, there is neither  a single transformation function that can model all patterns nor a single approach that can take advantage of all the different transformation functions.

In this paper, we tackle this problem and propose a general framework to integrate different transformation functions from several KGE models, $\mathbb{M}$, in a low-dimensional geometric space such that heterogeneous relational and structural patterns are well represented.
In particular, we employ spherical geometry to unify different existing representations of KGE queries, $(h,r,?)$.
In our framework, representations of KGE queries, $g_r^i(\textbf{h})$ with $i \in \mathbb{M}$, define the centers of hyperspheres, and candidate answers lie inside or outside of the hyperspheres whose radiuses are derived during training. 
Plausible answers mostly lie inside the convex hull formed by the centers of the hyperspheres. 
Based on this representation, we learn how to pay attention to the most suitable representations of a KGE query.
Thereby, attention is  acquired to adhere to applicable patterns (see \autoref{fig:patterns} and \autoref{MotivExamp}). 

For instance, given a KGE query $(\mathit{Leader1}, \mathit{coAuthor}, ?)$, attention will focus on the representation of this query defined by RotatE, as our framework has learned about the symmetry of the relation \emph{coAuthor}.
Likewise, TransE and RotatE will be preferred for \break KGE query $(\mathit{Student1}, \mathit{supervisedBy}, ?)$ accounting for the pattern\break $(X, \mathit{isPhDIn}, Y), (Y, \mathit{ledBy}, Z) \rightarrow (X, \mathit{supervisedBy}, Y)$, while TransE will be favored for KGE query  \emph{(Leader1, supervisedBy, ?)} due to the anti-symmetry of \emph{supervisedBy}.

Furthermore, we also project our model onto a non-Euclidean manifold, the Poincaré ball, to also facilitate structural preservation.

In summary, our key contributions are as follows:

\begin{itemize}
    \item We propose a spherical 
    geometric framework for combining several existing KGE models. 
    To our knowledge, this is the first  approach to integrate KGE models taking advantage of the different underlying geometric transformations.
    
    \item We utilize an attention mechanism to focus on query representations depending on the characteristics of the underlying relation in the query.
    Therefore, our method can support various relational patterns. Furthermore, structural patterns are captured by projecting the model onto the Poincaré ball.
    \item We present various theoretical analyses to show that our model subsumes various existing models.
\end{itemize}

%%% Local Variables:
%%% TeX-master: "main.tex"
%%% ispell-local-dictionary: "american"
%%% End:

\begin{table*}[t]
\centering
\caption{Specification of query representation of baseline and state-of-the-art KGE models and respective pattern modeling and inference abilities. AttE/H include both rotation (RotatE) and reflection (RefH), hence are not mentioned in the table to avoid repetitions. 
$\circ$ is element-wise complex product together with relation normalization.} 
\begin{tabular}{llllllll}

\toprule
Model & Query & Embeddings & Symmetry & Antisymmetry & Inversion & Composition & Hierarchy\\ 
\midrule
TransE \cite{bordes2013translating}  & $\boldsymbol{q = h + r}$ & $ \, \boldsymbol{q,h,r} \in \mathbb{R}^d$ & $\boldsymbol{\times}$ & $\boldsymbol{\checkmark -0}$ & $\boldsymbol{\checkmark -0}$ & $\boldsymbol{\checkmark -0}$ & $\boldsymbol{\times}$\\

RotatE \cite{sun2019rotate}   &   $\boldsymbol{q = h \circ r}$ & $ \, \boldsymbol{q,h,r} \in \mathbb{C}^d$ & $\boldsymbol{\checkmark -2}$ & $\boldsymbol{\checkmark -2}$ & $\boldsymbol{\checkmark -2}$ & $\boldsymbol{\checkmark -2}$ & $\boldsymbol{\times}$\\ 

ComplEx \cite{trouillon2016complex} & $\boldsymbol{q = h \times r}$ & $ \, \boldsymbol{q,h,r} \in \mathbb{C}^d$ & $\boldsymbol{\checkmark -2}$ & $\boldsymbol{\checkmark -2}$ & $\boldsymbol{\checkmark -2}$ & $\boldsymbol{\times}$ & $\boldsymbol{\times}$\\ 

DistMult \cite{Distmultyang2014embedding} & $\boldsymbol{q = h \cdot r}$ & $\, \boldsymbol{q,h,r} \in \mathbb{R}^d$ & $\boldsymbol{\checkmark -0}$ & $\boldsymbol{\times}$ & $\boldsymbol{\times}$ & $\boldsymbol{\times}$ & $\boldsymbol{\times}$\\ 

RefH \cite{chami2020low} & $\boldsymbol{q = Ref(\theta_r) h}$ & $\, \boldsymbol{q,h} \in \mathbb{H}^d$ & $\boldsymbol{\checkmark -0}$ & $\boldsymbol{\times}$ & $\boldsymbol{\times}$ & $\boldsymbol{\times}$ & $\boldsymbol{\checkmark -0}$\\ 

\bottomrule
\end{tabular}
\label{tbl:baselines}
\end{table*}
\section{Related work}
We review the related works in three parts, namely the baseline models we used for combination, the models which provide other approaches for combinations, and models that combine spaces.
\subsection{KGE Model Baselines}
Various models  \cite{DBLP:journals/semweb/GeseseBAS21,surveywang2017knowledge,ji2021survey}  have been proposed for KGE in the last few years. 
Each KGE defines a score function $f(h,r,t)$ which takes embedding vectors of a triple ($\boldsymbol{h,r,t}$) and scores the triple. In our work we integrate and compare them to the following baselines:

\begin{itemize}
    \item TransE \cite{bordes2013translating} computes the score of a triple by computing the distance between the tail and the \textit{translated} head.
    Thanks to the translation-based transformation, this KGE is particularly suited for modeling inverse and composition patterns.
    \item RotatE \cite{sun2019rotate} uses a relation-specific rotation $\mathbf{r}_i=e^{i \theta}$ to map each element of the head to the corresponding tail. RotatE can infer symmetrical patterns if the angle formed by the head and tail is either 0 or $\theta$. Besides, rotations are also effective in capturing antisymmetry, composition, or inversion.
    
    \item DistMult \cite{Distmultyang2014embedding} represents each relation as a diagonal matrix. Its score function captures pairwise interaction between \textit{the same dimension} of the head and tail embedding. Thus, DistMult treats symmetric relations well, but scores so highly inverse links of non-symmetric and antisymmetric relations. 
    \item ComplEx \cite{trouillon2016complex} extends DistMult in the complex space to effectively capture symmetric and antisymmetric patterns.
    \item AttH \cite{chami2020low} combines relation-specific rotations and reflections using hyperbolic attention and applies a hyperbolic translation. Rotation can capture antisymmetrical and symmetrical patterns, reflection can naturally represent symmetrical relations, while the hyperbolic translation can capture hierarchy. We also compared our models against AttE \cite{chami2020low}, a variant of AttH with curvature set to zero. 
\end{itemize}

\subsection{KGEs Combination}
\paragraph{Combinations between KGEs of the same kind}
Authors in \cite{xu2021multiple} showed that, under some conditions, the ensemble generated from the combination of multiple runs of low-dimensional embedding models \textit{of the same kind} outperforms the corresponding individual high-dimensional embedding model. Unlike our approach, the ensemble model will still be able to express only a subset of existing logical patterns.

\paragraph{Combination between different KGE models}

Prior works \cite{krompass2015ensemble} proposed to combine different knowledge graph embeddings through score concatenation to improve the performance in link prediction. \cite{wang2018multi} proposed a relation-level ensemble, where the combination of individual models is performed separately for each relation. A recent work \cite{wang2022probabilistic}, proposed to combine the scores of different embedding models by using a weighted sum. Such methods combine scores either per model or per relation, while we provide a query attention mechanism for the combination. 

A different approach has been proposed in MulDE \cite{wang2021mulde}, where link prediction is improved by \textit{correcting} the prediction of a ``student'' embedding through the use of several pre-trained embeddings that act as ``teachers''. The student embedding can be considered to constitute an ensemble model. However, this ensemble cannot steer decisions towards the strengths of individual models but can only decide randomly or based on majority guidance by teachers.

Further ensemble approaches between KGE and machine learning models can be found in the Appendix \ref{rel_work_app}

\subsection{Combination Of Spaces}
A different line of research aims at improving link prediction performance by combining different geometrical spaces. \cite{gu2018learning} improves link prediction by combining Hyperbolic, Spherical, and Euclidean space. 
Similarly, \cite{xiong2022ultrahyperbolic} embedded knowledge graphs into an Ultra-hyperbolic manifold, which generalizes Hyperbolic and Spherical manifolds. On the other hand, we combine queries rather than geometric spaces.

\section{Proposed approach}

In this section, we present our geometric query integration model using Euclidean and Hyperbolic geometries, and introduce our approach in the following four items:
a) entity, relation, and query representation,  
b) spherical query embedding, 
c) Riemannian attention-based query combination, 
d) expressivity analysis.

\begin{figure}[!ht]
    \centering
    \includegraphics[width=\hsize]{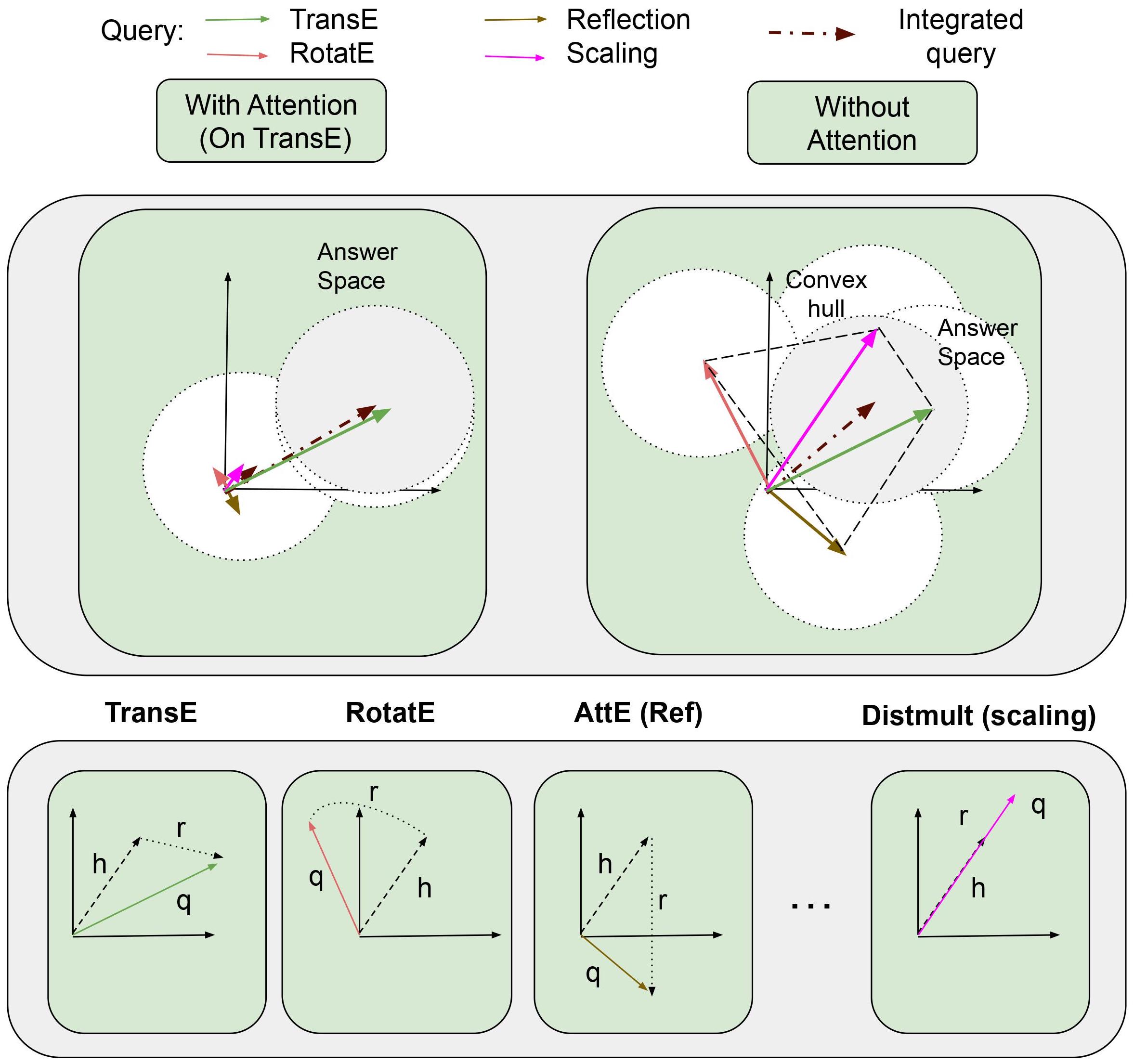}
    \caption{The overall architecture of our proposed model with spherical geometry. We combine query representations of TransE, RotatE, AttE (with Reflection), and DistMult (per dimension scaling). The left part shows query integration with attention to TransE model. The right part represents query combination without attention.}
    \label{MotivExamp}
\end{figure}
\paragraph{\textbf{a) Entity, Relation and Query Embedding}}
\label{method:representation}
Let $\mathcal{E}, \mathcal{R}$ be the entity and relation sets. 
We represent each entity $e \in \mathcal{E}$ and relation $r \in \mathcal{R}$ as $d_e$ and $d_r$ dimensional vectors which are denoted by $\boldsymbol{e}$ and $\boldsymbol{r}$, respectively. 
Thus, each triple $(h,r,t)$ has a vector representation  $(\boldsymbol{h,r,t})$ where $\boldsymbol{h,t}$ are the corresponding entity embeddings.

We split each triple $(h,r,t)$ into two parts, namely the tail query $q= (h,r,?)$ and the candidate answer $t$, and represent their embeddings by $\boldsymbol{q,t}$ respectively.

In our model, we aim at combining the queries from several existing KGE models that are specified in \autoref{tbl:baselines}. 
We denote the query representation set by 
$\mathcal{Q} = \{ \boldsymbol{q}_i | \boldsymbol{q}_i = g_r^i(\boldsymbol{h}) , i \in \mathbb{M}\}$ where $\mathbb{M}$ is a set of several existing KGE models such as TransE, RotatE, ComplEx, DistMult, etc, and the function $g_r^i(\boldsymbol{h})$ is a relation-specific transformation from a head embedding to a query representation for model $i$.
Note that we assume that query representations by different models lie on the same space. In this paper, we stay in Euclidean space for query combination. In this regard, we can combine models lying either directly in Euclidean space (e.g., TransE and DistMult) and models that can be rewritten to lie in the Euclidean space (e.g., models lying in Complex or Hypercomplex spaces as ComplEx, RotatE, and QuatE by assuming $\mathbb{R}^4 = \mathbb{R}^2 \times \mathbb{R}^2 = \mathbb{C}^1 \times \mathbb{C}^1$, where $\mathbb{C}^d, \mathbb{R}^d$ are $d$-dimensional Complex and Euclidean spaces). We then project such query vectors on a hyperbolic manifold to handle hierarchical patterns.

\paragraph{\textbf{b) Spherical Query Embedding}}
\label{sphericalellipticalquery}

In this part, first, we propose a spherical query embedding to represent each query as a sphere whose center is the vector embedding of the query. This sphere
defines the answer space of the query.
Second, we propose an approach to combine query representations of several already existing embedding models in one spherical query representation to enhance the modeling of heterogeneous patterns.  
In ``\textit{radius and ranking}", we will show that the spherical representation is connected to the ranking metric Hits@k.
In particular, the top k candidate answers for a query $q$ are embedded in a sphere whose center is a combination of the vector embeddings $\boldsymbol{q}_i$ of query $q$.
To practically enforce this, the radius in our spherical query embedding needs to be set. 
Therefore, in  ``\textit{radius and loss}", 
we will show that a loss function can enforce the improvement of Hits@k by enforcing top k candidate answers of a query inside the sphere.

Here, we formalize the combination of $n$ spherical KGEs. 
Let us assume that $\boldsymbol{q_1, q_2, \ldots, q_n} \in \mathcal{Q}$ be the $n$ vector query embeddings of a query $q=(h,r,?)$ from $n$ distinct KGE models, and $\boldsymbol{a = t}$ be the embedding of the candidate answer.
We represent each query as a hypersphere with a pair $\boldsymbol{q}^c_i = (\boldsymbol{q_i}, {\epsilon_{i}}), \,\, \boldsymbol{q}_i \in \mathcal{Q}$,
where $\boldsymbol{q_i} \in \mathbb{R}^d$ is the center of the $i$th sphere associated to the $i$th model and ${\epsilon}_{i}$ is the radius. 
By using the function
\begin{equation}
    p(\boldsymbol{q}_i, \boldsymbol{a}) = \| \boldsymbol{a} - \boldsymbol{q_i} \|,\quad
    \boldsymbol{q}_i \in \mathcal{Q},
\end{equation}
we define the answer space $\mathcal{A}$ and non-answer space $\mathcal{N}$ as decision boundaries in the embedding space for each query as follows:
\begin{equation}
    \begin{cases}
        \mathcal{A}_i &= \{ \boldsymbol{e} \in \mathbb{R}^d \,\,\, | \,\,\, \| \boldsymbol{e} - \boldsymbol{q}_i \| \leq \epsilon_{i}\},
        \\
        \mathcal{N}_i &= \{ \boldsymbol{e}  \in \mathbb{R}^d \,\,\, | \,\,\, \| \boldsymbol{e} - \boldsymbol{q}_i \| > \epsilon_{i}\}.
    \end{cases}
\end{equation}
In this case, all embeddings of answers $a$ are supposed to lie on or inside a sphere with a radius of $\epsilon_{i}$ and center $\boldsymbol{q}_i$, i.e., $\boldsymbol{a} \in \mathcal{A}_i$, and the ones which are not answers lie outside of the sphere \cite{nayyeri2021loss,zhou2017learninglimit}.
We combine spherical query embeddings of several existing KGE models in one spherical query embedding as follows:

\paragraph{Combination}

Given the vector embeddings $\boldsymbol{q_1, q_2, \ldots, q_n} \in \mathcal{Q}$ 
we can set a radius $\epsilon^k_i$ for each $\boldsymbol{q_i}$ such that the answer space $\mathcal{A}_i$ covers the top $k$ candidate answers.
\begin{equation}
    \begin{cases}
    \mathcal{A}_1 &= \{ \boldsymbol{e} \in \mathbb{R}^d \,\,\, | \,\,\, \| \boldsymbol{e} - \boldsymbol{q}_1 \| \leq \epsilon_1^k\},
    \\
    &\vdots
    \\
   \mathcal{A}_n &= \{ \boldsymbol{e} \in \mathbb{R}^d \,\,\, | \,\,\, \| \boldsymbol{e} - \boldsymbol{q}_n \| \leq \epsilon_n^k\}.
    \end{cases}
\end{equation}
Summing up the above inequalities, we have
$\| \mathbf{a} - \mathbf{q}_1\| + \ldots + \|\mathbf{a} - \mathbf{q}_n\| \leq  \epsilon_1^k + \ldots + \epsilon_n^k.$
Because of triangle inequality of the metric $||.||$, this can be extended to the following inequality
$\| \mathbf{a} - \mathbf{q}_1 + \ldots + \mathbf{a} - \mathbf{q}_n\| \leq  \epsilon_1^k + \ldots + \epsilon_n^k,$ 
that concludes 
$\| \mathbf{a} - \frac{\mathbf{q}_1 + \ldots + \mathbf{q}_n}{n} \| \leq  \frac{\epsilon_1^k + \ldots + \epsilon_n^k}{n}.$
Therefore, the combined spherical query embedding is the spherical embedding $\boldsymbol{q}_E^c = (\boldsymbol{q}_E, \epsilon_{E})$ where 
\begin{equation}
    \begin{cases}
    \boldsymbol{q}_E &= \frac{\boldsymbol{q}_1 + \ldots + \boldsymbol{q}_n}{n},
    \\
    \epsilon_{E} &= \frac{\epsilon_1^k + \ldots + \epsilon_n^k}{n}.
    \end{cases}
    \label{eq:simpleEnsemble}
\end{equation}
This leads to the following top $k$ candidate answer space of the combined spherical query embedding:
\begin{equation}
    \mathcal{A}_E = \{ \boldsymbol{e} \in \mathbb{R}^d \,\,\, | \,\,\, \| \boldsymbol{e} - \boldsymbol{q}_E \| \leq \epsilon_{E}\}.
    \label{combinedquery}
\end{equation}
\autoref{MotivExamp} (top right) shows the query representations, and candidate answer spaces of TransE, RotatE, RefE, and DistMult, together with the combined query (without attention to a particular model). 
The combined query mainly lies inside the convex hull of all the models within the answer space. 
We later show that most answers lie within the convex hull covered by the combined query. Therefore, the combined model takes the advantage of all models. 
Before theoretically justifying this, we bridge the radius $\epsilon$ in spherical query embedding and ranking metrics, as well as the practical way of modeling radius using the loss function in the following parts.

\paragraph{Radius and Ranking}
Most KGE models are evaluated based on ranking metrics such as Hits@k \cite{bordes2013translating}.
Here we explain the connection between the ranking metrics and radius in our spherical query embedding.
Because the overall ranking is computed by taking the average over ranks of all test triples, 
we explain the connection between ranking and our model by considering an individual test triple.
During testing, for each given positive test triple $(h,r,t)$, the tail $t$ is replaced one by one with all entities $e\in \mathcal{E}$.
We denote $\mathbb{T}_e = (h,r,e)$ the corrupted triple generated by replacing $t$ by $e$.
Therefore, $\mathcal{T} = \{\mathbb{T}_e | e \in \mathcal{E} - \{t\}\}$ is the set of all corrupted triples generated from the correct triple $(h,r,t)$. 
After computing the score of each triple in $\mathcal{T}$ and sorting them based on their scores in descending way, 
we select top $k$ high score samples and generate a new set $\mathcal{T}_k$ containing these samples.
 The spherical query embedding $\boldsymbol{q}_E^c=(\boldsymbol{q}_E,\epsilon_E)$ associated to a query $q=(h,r,?)$ defines a spherical answer space $\mathcal{A}_E$ that contains the vector embeddings $\boldsymbol{e}$ for the top $k$ entities $e \in\mathcal{T}_k$. That is,
$\mathcal{T}_k$ contains top $k$ candidates for a query $q$, and $\mathcal{A}_E$ in \autoref{combinedquery} is the candidate answer embedding space. 
We want the vectors of answers in $\mathcal{T}_k$ to lie inside $\mathcal{A}_E$, and to be as close as possible to the query center to improve ranking results.
To enforce this, we define a loss function to optimize the embeddings, as is explained below.

\paragraph{Radius and Loss Function}
In this part, we show that the existing loss functions implicitly enforce a particular (implicit) radius around the vector query embedding $\boldsymbol{q}_E$. 
Let us focus on the widely used loss function shown in the following \cite{chami2020low}:
\begin{equation}
\mathcal{L} = \sum_{e \in \mathcal{E}} \log(1 + \exp(y_{e}  (-p(\boldsymbol{q}_i, \boldsymbol{e}) \\
    + \delta_{h} + \delta_{e})))),
\end{equation}
where $y_{e} = 1$ if $e = a$, and $y_{e} = -1$ if $e \neq a$, and $\delta_{h}$ and $\delta_{e}$ are trainable entity biases.
Minimization of this loss function leads to maximizing the function $-p(\boldsymbol{q}_i, \boldsymbol{e}) + \delta_{h} + \delta_{e}$. 
This can be approximately represented as  
$-p(\boldsymbol{q}_i, \boldsymbol{e}) + \delta_{h} + \delta_{e} \geq M$ where $M$ is a large number. 
Therefore, we have $p(\boldsymbol{q}_i, \boldsymbol{e}) \leq \delta_{h} + \delta_{e} - M = \delta_{he} - M = \epsilon_i$ which forms boundaries for classification as well as ranking. 
In the next part, we theoretically show that $\boldsymbol{q}_E$ lies  within the convex hull of the set of vectors $\{\boldsymbol{q}_1,\dots,\boldsymbol{q}_n\}$.
Thus, the combined model takes advantage of each model in ranking.

\paragraph{Theoretical Analysis}
\autoref{combinedquery} indicates that if the query is represented by $(\boldsymbol{q}_E, \epsilon_{E})$, then the score given by the combined model to a plausible answer is lower than the average of the scores given by the individual models, and higher than the lowest individual model score because, without loss of generality, we have
\begin{equation}
    \min(p(\boldsymbol{q}_1, \boldsymbol{e}), p(\boldsymbol{q}_2, \boldsymbol{e})) \leq p(\boldsymbol{q}_E, \boldsymbol{e}) \leq \max(p(\boldsymbol{q}_1, \boldsymbol{e}), p(\boldsymbol{q}_2, \boldsymbol{e})).
\end{equation}
\noindent
This equation shows that for a particular $k$, the combined model gets a better score than the worst model, but it gets a lower score than the best one. 
However, by increasing $k$, the combined model covers the answers provided by both models because most of the answers lie in the convex hull between the queries (later it will be proved), and the combined model with arbitrary large $k$ covers the answers represented by each model. 
Therefore, the combined model improves Hits@k with a sufficiently large $k$. Later in this section, we present the attention-based model which enables us to improve Hits@k for small $k$.

The following proposition states that the best embedding for an answer to a query lies in the convex hull of the query embeddings given by two models. This implies that if two models are trained jointly with the combined model, the answers of each query lie between the centers of the two spheres associated with the two embeddings of the query. This facilitates the answer space of combined spherical query embedding to cover the answer embedding from each individual model.
This can be generalized for an arbitrary number of models.

\begin{proposition}
\label{proposition:spherical:answerbetween}
Let $\boldsymbol{q}_1$ and $\boldsymbol{q}_2$ be two query embeddings for a query $q$. Then, the following two statements are equivalent for every vector $\boldsymbol{a}$ in the vector space:
\begin{align}
  \left\{
   \begin{aligned}
    & \boldsymbol{a} = \operatorname{argmin}_{\boldsymbol{e}} (p(\boldsymbol{q}_1, \boldsymbol{e}) + p(\boldsymbol{q}_2, \boldsymbol{e})), \\
    &\boldsymbol{a} \text{ lies in the convex hull of vectors } \boldsymbol{q}_1 \text{ and } \boldsymbol{q}_2.
    \end{aligned}
    \right.
\end{align}
\end{proposition}

\paragraph{\textbf{c) Riemannian Attention-Based Query Combination.}}
\paragraph{Weighted Combined Query Embedding}
A consequence of proposition \ref{proposition:spherical:answerbetween} is that the combined query embedding can improve the performance when $k$ is sufficiently large (e.g., Hits@20).
However, for a low $k$ (e.g., Hits@1) the performance is degraded because one model gets a better ranking, and the combined model with an average query does not cover it. 
In addition, among several models, there might be possible that some models return wrong answers which might also influence the combined model.
Therefore, allowing the combined spherical query embedding $\boldsymbol{q}_E$ to slide to $\boldsymbol{q}_1$ or $\boldsymbol{q}_2$ is beneficial.
Hence, without loss of generality, we combine two query embeddings as the convex combination of the inequalities:
\begin{equation}
    \begin{cases}
    \alpha \| \boldsymbol{a} - \boldsymbol{q}_1\| \leq \alpha \epsilon_1^k, 
    \\
    \beta \|\boldsymbol{a} - \boldsymbol{q}_2\| \leq \beta \epsilon_2^k,\quad
    \alpha, \beta \geq 0,\;
    \alpha + \beta = 1.
    \end{cases}
\end{equation}

By computing this convex combination, we have
$ \alpha \| \boldsymbol{a} - \boldsymbol{q}_1\| + \beta \|\boldsymbol{a} - \boldsymbol{q}_2\| \leq  \alpha \epsilon_1^k + \beta \epsilon_2^k.$ This inequality implies
$ \| \alpha \boldsymbol{a} -  \alpha \boldsymbol{q}_1  + \beta  \boldsymbol{a} -  \beta \boldsymbol{q}_2\|  \leq \alpha \| \boldsymbol{a} - \boldsymbol{q}_1\| + \beta \|\boldsymbol{a} - \boldsymbol{q}_2\| \leq  \alpha \epsilon_1^k + \beta \epsilon_2^k,$ 
which subsequently leads to
\begin{equation}
    \| \boldsymbol{a} -  (\alpha \boldsymbol{q}_1  +  \beta \boldsymbol{q}_2) \| \leq  \alpha \epsilon_1^k + \beta \epsilon_2^k.    
    \label{eq:weightedEnsembleSpherical}
\end{equation}

Therefore, the combined spherical query embedding is $\boldsymbol{q}_E^c=(\boldsymbol{q}_E, \epsilon_E^k)$ where $\boldsymbol{q}_E = (\alpha \boldsymbol{q}_1 + \beta \boldsymbol{q}_2)$ and $\epsilon_{E}^k = \alpha \epsilon_1^k + \beta \epsilon_2^k.$ 
This combination is generalized for $n$ models:
\begin{equation}
    \textstyle
    \| \boldsymbol{a} -  \sum{\alpha_i \boldsymbol{q}_i} \| \leq  \sum{\alpha_i \epsilon_i^k},
\end{equation}

\paragraph{Attention Calculation.}

Given a combined spherical query embedding $\boldsymbol{q}_E^c=(\boldsymbol{q}_E, \epsilon_E)$ with 
\begin{equation}
    \boldsymbol{q_E} = \sum \alpha_i \mathbf{q}_i, \epsilon_{E}^k = \sum \alpha_i \epsilon_i^k,
    \label{eq:attentionquery}
\end{equation}
we can compute $\alpha_i$s by providing an attention mechanism \cite{chami2020low}
\begin{equation}
\alpha_i = \frac{\exp(g(\boldsymbol{w}\boldsymbol{q}_i))}{\sum_j \exp(g(\boldsymbol{w}\boldsymbol{q}_j))}, 
\label{eq:attentioncompute}
\end{equation}
where $g(\boldsymbol{x}) = \boldsymbol{w x}$ is a function with a trainable parameter $\boldsymbol{w}$. We call this version of our model Spherical Embedding with Attention~(SEA).

\paragraph{Riemannian Query Combination.}

We next extend our attention-based query combination to Riemannian manifolds to model both relational patterns (via various transformations used in different models) and structural patterns as hierarchy via the manifolds (e.g., Poincaré ball). 
Similarly to \cite{chami2020low}, we perform attention on tangent space. We consider all models in Euclidean space and combine their query embeddings.
The resulting query embedding on the tangent space is then projected to the manifold via the exponential map.
This attention-based model combination is defined as follows:
\begin{equation}
\begin{cases}
    \boldsymbol{q}^{euc}_E &= \sum_i \frac{\exp(g(\boldsymbol{q}_i))}{\sum_j \exp(g(\boldsymbol{q}_j))} \boldsymbol{q}_i, \\
    \boldsymbol{q}^M_E &= \exp_{\boldsymbol{0}}(\boldsymbol{q}^{euc}_E).
\end{cases}
\end{equation}

We compute the score as 
$p(q, a) = d(\boldsymbol{q}^M_E \oplus  \boldsymbol{r}, \boldsymbol{a})$
, where $\boldsymbol{h,r,t,q}$ are points on a manifold $\mathcal{M}$, $\exp_0(\cdot)$ is the exponential map from origin, and $\oplus$ is Möbius addition.
In terms of the Poincaré ball, the manifold, exponential map, and Möbius addition are defined as follows \cite{balazevic2019multi, chami2020low}:
\begin{equation}
    \begin{cases}
        \mathcal{M} &= \{ \boldsymbol{p} \in \mathbb{R}^d | \| \boldsymbol{p} \| \leq \frac{1}{c} \}, \\
        \exp_{\boldsymbol{0}}(v) &= \tanh(\sqrt{c} \|\boldsymbol{v}\|) \frac{\boldsymbol{v}}{\sqrt{c} \|\boldsymbol{v}\|}, \\
        d^c(\boldsymbol{p,q}) &= \frac{2}{c} \tanh^{-1}( \sqrt{c} \| -\boldsymbol{p} \oplus \boldsymbol{q} \|), \\
        \boldsymbol{p} \oplus \boldsymbol{q} &= \frac{( 1 + 2c \langle \boldsymbol{p} , \boldsymbol{q} \langle + c \| \boldsymbol{q} \|^2) \boldsymbol{p} + ( 1 - c \|\boldsymbol{p}\|^2 ) \boldsymbol{q}}{1 + 2c \langle \boldsymbol{p} , \boldsymbol{q} \rangle + c^2 \|\boldsymbol{p}\|^2 \|\boldsymbol{q}\|^2},
    \end{cases}
\end{equation}
\noindent
where $c$ is the curvature, $\exp$ is the exponential map from a point on tangent space to the manifold, $d^c$ is the distance function with curvature $c$, and $\boldsymbol{v}$ is the point on the tangent space to be mapped to manifold via the exponential map. We call the hyperbolic version of our model Spherical Embedding with Poincaré Attention (SEPA).

\paragraph{\textbf{d) Expressivity Analysis}}
In this section, we analyze our models in terms of expressive power as well as the subsumption of other models. Our model is a generalization of various existing Euclidean and non-Euclidean KGE models. We say that a model $m_1$ subsumes a model $m_2$ if for every given KG $G$ and scoring by model $m_2$, there exists a scoring by model $m_1$ such that the score of every triple $t \in G$ by $m_1$ approximates the score of $t$ by $m_2$~\cite{kazemi2018simple, nayyeri20205}.

\begin{proposition}
SEPA subsumes AttH, and SEA subsumes TransE, RotatE, ComplEx, DistMult and AttE.
\label{subsumption}
\end{proposition}

\begin{corollary}
SEPA and SEA can infer anti-symmetry, symmetry, composition, and inversion patterns.
\label{inferpatterns}
\end{corollary}

\noindent
It is important to notice that a model can infer a pattern inherently or infer a pattern under a certain condition (see \autoref{tbl:baselines}). 
Our model aims to take advantage of the inference power of multiple models on heterogeneous patterns with minimum certain conditions by providing attention mechanisms per relation type forming different patterns.
Note that our model is not influenced by incapable models on particular patterns because the attention can be learned as zero for those models.
Overall, our combined model can inherit the capabilities mentioned in \autoref{tbl:baselines} and ignore the incapability of other models which is shown in \autoref{subsumption} and \autoref{inferpatterns}.
 Hence, if our model is executed on a dataset containing only a single pattern, we do not expect to outperform the combined models, rather than achieving competitive performance to the best model.

Proof of propositions can be found in Appendix \ref{proof_appendix}.

\section{Experiments}
In this section, we conduct extensive evaluations to show the effectiveness of our proposed approach. To do so, we first introduce the utilized datasets, 
followed by the selected baseline for the combination and the comparison.
We then present the experimental setup and hyper-parameter setting.
The results and analysis are presented in three folds: comparison with the individual baselines, comparison with other combination models, and comparison with models in the Ultrahyperbolic space.
Finally, we provide several analyses to show the role of attention on learning and inference over various patterns for different kinds of relations and models.

\begin{figure}[t]
  \includegraphics[scale=0.48]{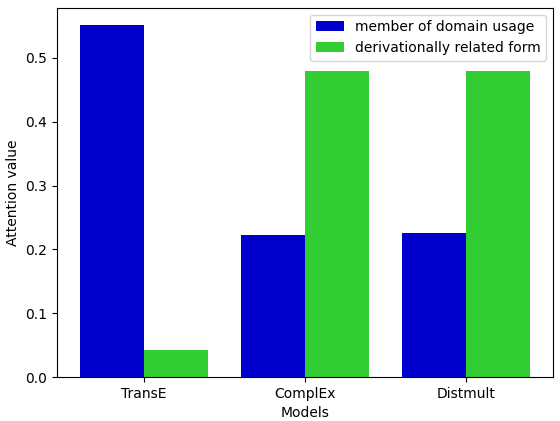}
  \centering
  \caption{Comparison between the importance given by each model to a symmetric (in green) and antisymmetric (in blue) relation.\protect\footnotemark}
  \label{plot_rel}
\end{figure}

\subsection{Dataset}
\footnotetext{E.g.\ \emph{ethics} is a form that is derivationally related to \emph{ethicist}.}We use the following standard benchmark for the evaluation:
\begin{itemize}
    \item \textbf{Wordnet}: WN18RR \cite{dettmers2018convolutional} is a subset of WN18, which contains a mixture of symmetric and antisymmetric as relational patterns, and  hierarchical structural patterns. \textit{see also} and \textit{hypernym} are examples of symmetry and hierarchy in this dataset. WN18RR contains 11 relations, 86,835 training triples, and 40,943 entities. Compared to the other datasets in KGE literature, WN18RR is considered sparse;
    \item \textbf{FreeBase}: FB15k-237 \cite{toutanova2015observed} is the subset of FB15k from removing leakage of inverse relations \cite{dettmers2018convolutional}. 
    FB15k-237 is less sparse than WN18RR, and mainly contains composition patterns. It contains 237 relations, 272,115 triples, and 14,541 entities.
    \item \textbf{NELL}: NELL-995 \cite{xiong2017deeppath} contains 75,492 entities and 200 relations, having $\sim$ 22\% hierarchical relations. We use a subset of NELL-995 with 100\% hierarchy, created in \cite{balazevic2019multi}. 
\end{itemize}

\subsection{Baseline}
In this section, we aim to show experimentally that the geometric combination of several existing 
KGE models improve their performance.
 To this end, we select a subset of KGEs in Euclidean, Complex, and Hyperbolic space with different capabilities to show we can combine a wide range of models. In particular, we select a subset of TransE, DistMult, ComplEx, RotatE, AttE (only reflection), and AttH (hyperbolic projection operator) and compare our combined models against such baselines. We also compare our models with two additional state-of-the-art KGEs: in high dimension, TuckER \cite{DBLP:conf/emnlp/BalazevicAH19}, and in low dimension, MuRP \cite{balazevic2019multi}, to show that our models can outperform models that were not combined.
Furthermore, we also compare our model with a recent top model for combining several KGEs, namely MulDE \cite{wang2021mulde} because it uses a similar set of KGEs for the combination, similar dimensions, and some of the benchmarks we used.
Additionally, we will show that our model gets comparable performance with UltraE \cite{xiong2022ultrahyperbolic}, a model on the Ultrahyperbolic space.

\subsection{Experimental Setup}
\paragraph{\textbf{Evaluation Metrics}}
We use the popular ranking metrics \cite{surveywang2017knowledge} namely Mean Reciprocal Rank (MRR), and Hits@k, k = 1,3,10. Given a set of test triples $\mathcal{T} = \{(h,r,t)\}$, for each test triple $p = (h,r,t)$, we compute its rank as follows:
we first corrupt the head entity by replacement with all possible entities in the KG, say $e' \in \mathcal{E}$, and generate a set of candidate corrupted triples for $p$ i.e., $p_c = \{p' = (e',r,t)\}$.
We filter $p_c$ by removing all generated candidates which are already appeared in the train, validation, and test sets, together with removing the cycle. After computing the score of the candidate triples and sorting them, we find the rank of the candidate test $p$, and call it $r_p$. The same procedure is performed for computing the right rank by corrupting the tail entity and computing the right rank. The average of the left and right ranks will be considered the final rank of the test triple.
We then compute the average reciprocal of all test triples and report it as MRR. Hits@k will be computed by reporting the percentage of the test triples ranked less than $k$.

\paragraph{\textbf{Hyperparameters}}
The hyperparameters corresponding to our model are embedding dimension $d$, models to combine $m$, optimizer $o$, learning rate $lr$, number of negative samples $n$, batch size $b$, dtype $dt$, and double\_neg $dn$.
We additionally used $\alpha^2$ as attention parameters (in place of $\alpha$) playing the role of a simple kind of regularization mechanism ($ar$), to further penalize the models with less contribution in the attention. 
Following the common practice of KGEs, we use both low $d=32$ and high dimensions $d=500$ for the evaluation of our model. For the other hyperparameters, we use the following ranges $m=$ \{TransE, DistMult, ComplEx, RotatE, AttE (only reflection)\},
$o =$ \{Adam, Adagrad\}, $lr = \{0.1, 0.05, 0.001\}$, $n = \{-1, 50, 100, 150, 200, 250\}$ where -1 refers to full negative sampling \cite{lacroix2018canonical}, $b = \{ 100,500\}$, $dt = \{ single, double\}$, $ar=\{yes,no\}$, $dn=\{yes,no\}$. We also add reciprocal triples to the training set as the standard data augmentation technique \cite{lacroix2018canonical}. The optimal hyperparameters for each dataset are specified in Appendix \ref{hyperparam_appendix}.

\begin{table}[h]
\centering
\small
\caption{Comparison of H@1 for WN18RR relations. TE = TransE, CE = ComplEx, DM = DisMult}
\begin{tabular}{l|llll}
\hline
\multirow{1}{*}{Relation} & \multirow{1}{*}{TE}& \multirow{1}{*}{CE}& \multirow{1}{*}{DM} & \multirow{1}{*}{SEPA}
\\\hline
member meronym & \underline{0.144} & 0.095 & 0.030 & \textbf{0.162}\\
hypernym & 0.060 & \underline{0.064} & 0.024 & \textbf{0.121}\\
has part & \underline{0.105} & 0.099 & 0.029 & \textbf{0.125}\\
instance hypernym & \textbf{0.246} & 0.242 & 0.139 & 0.242\\
member of domain region & \underline{0.269} & 0.077 & 0.096 & \textbf{0.327}\\
\underline{member of domain usage} & \underline{\textbf{0.271}} & 0.188 & 0.062 & \textbf{0.271}\\
synset domain topic of & \underline{0.289} & 0.136 & 0.070 & \textbf{0.329}\\
also see & 0.161 & 0.580 & \textbf{0.598} & 0.571\\
\underline{derivationally related form} & 0.532 & \underline{0.943} & 0.940 & \textbf{0.944}\\
similar to & 0.000 & \underline{\textbf{1.000}} & \underline{\textbf{1.000}} & \textbf{1.000}\\
verb group & 0.333 & 0.949 & \textbf{0.974} & 0.923\\
\hline

\end{tabular}
\label{tbl:comp_rel}
\end{table}

\subsection{Link Prediction Results And Analysis}
The result of comparing SEA and SEPA to the combined models on FB15k-237, WN18RR and NELL-995-h100 are shown in Table \ref{tab:main_table_low} ($d = 32$) and in Table \ref{tab:main_table_high} ($d = 500$). 

As expected, while the hyperbolic version of our combined model (SEPA) outperforms all baselines in low-dimensional settings, the Euclidean one (SEA) is the best model in high-dimensional space. Comparing SEPA and SEA, in low-dimensional space, we can see the performance improvements on WN18RR and NELL-995-h100 are much more than FB15k-237.
This is due to the presence of a significant amount of hierarchical relations in WordNet and NELL compared to Freebase. We still observe SEPA outperforms SEA on FB15k-237 dataset. The main reason is that SEPA combines hyperbolic manifolds with various transformations used in queries of different models, so it is capable of capturing the mixture of structural and logical patterns in a low dimension (e.g., compositional patterns in Freebase).
Even though we did not combine AttE and AttH directly, but only used reflection and the hyperbolic projection, respectively, we were still able to outperform them.
 Similarly, SEPA outperforms MuRP in low dimensions, and SEA outperforms TuckER in high dimensions in all metrics apart from the H@1 of FB15k-237.
More details are available in Appendix \ref{comp_tuck}.

Our combination model 
increases the expressiveness of individual models (Proposition \ref{subsumption}), having the best performance gain in low-dimensional space. Besides, 
our model takes advantage of the inference power of the base models with fewer constraints (Table \ref{tbl:baselines}) by utilizing the attention mechanism.
On the other hand, in high-dimensional space, Euclidean models are proven to be fully expressive \cite{wang2018multi}.
 Hence, even though SEA outperforms all baselines, the performance gain is not as significant as in low-dimensions. 

\begin{table*}[ht]
\centering
\small
\begin{adjustbox}{width=\textwidth,center}
\begin{tabular}{|l|l|cccc|cccc|cccc|}
\hline
\multirow{2}{*}{\textbf{Elements}} & \multirow{2}{*}{\textbf{Model}} & \multicolumn{4}{c|}{\textbf{WN18RR}} & \multicolumn{4}{c|}{\textbf{FB15k-237}} & \multicolumn{4}{c|}{\textbf{NELL-995-h100}} \\ \cline{3-14}
&  & MRR & H@1 & H@3 & H@10 &  MRR & H@1 & H@3 & H@10 &  MRR & H@1 & H@3 & H@10 \\ \hline
\multirow{3}{*}{Individual Models} 
&TransE & 0.358  & 0.263 & 0.423 & 0.527 & 0.292 & 0.208 & 0.319 & 0.461 & 0.267 & 0.188 & 0.298 & 0.423 \\
&DistMult & 0.383 & 0.370 & 0.385 & 0.405 &  0.299 & 0.209 & 0.330 & 0.477 & 0.279 & 0.207 & 0.308 & 0.419 \\
&RotatE & 0.389  & 0.376 & 0.392 & 0.410 & 0.258 & 0.178 & 0.284 & 0.416 & 0.264 & 0.193 & 0.292 & 0.405 \\
&ComplEx & 0.419  & 0.395 & 0.426  & 0.465 & 0.264 & 0.184 & 0.289 & 0.421 & 0.247 & 0.176 & 0.275 & 0.387 \\
&AttE & 0.463  & \underline{0.430} & 0.474 & 0.528 & \underline{0.314} & \underline{0.227} & 0.343 & 0.489 & \underline{0.334} & \underline{0.247} & \underline{0.375} & \underline{0.503}\\
&AttH & \underline{0.468} & 0.429 & \underline{0.485} & \underline{0.539} &  \underline{0.314} & 0.223 & \underline{0.346} & \underline{0.498} & 0.326 & 0.240 & 0.367 & 0.493 \\
\hline
\hline
\multirow{2}{*}{Our models} 
& SEPA & \textbf{0.481} & \textbf{0.441} & \textbf{0.496} & \textbf{0.562} &  \textbf{0.332} & \textbf{0.243} & \textbf{0.363} & \textbf{0.509} & \textbf{0.346} & \textbf{0.261} & \textbf{0.385} & \textbf{0.508}\\
& SEA & 0.468 & 0.430 & 0.485 & 0.538 & 0.326 & 0.238 & 0.356 &0.504 & 0.333 & 0.245 & 0.376 & 0.504 \\
\hline
\hline
\multirow{1}{*}{Ablation} 
&  SEP & 0.478 & 0.437 & 0.493 & 0.556 & 0.329 & 0.239 & 0.361 & \textbf{0.509} & 0.340 & 0.254 & 0.380 & 0.505 \\
\hline
\end{tabular}
\end{adjustbox}
\caption{Link prediction evaluation on datasets for d=32. Best score and best baseline are in bold and underlined, respectively. 
}
\label{tab:main_table_low}
\begin{adjustbox}{width=\textwidth,center}
\begin{tabular}{|l|l|cccc|cccc|cccc|}
\hline
\multirow{2}{*}{\textbf{Elements}} & \multirow{2}{*}{\textbf{Model}} & \multicolumn{4}{c|}{\textbf{WN18RR}} & \multicolumn{4}{c|}{\textbf{FB15k-237}} & \multicolumn{4}{c|}{\textbf{NELL-995-h100}} \\ \cline{3-14}
&  & MRR & H@1 & H@3 & H@10 &  MRR & H@1 & H@3 & H@10 &  MRR & H@1 & H@3 & H@10 \\ \hline
\multirow{3}{*}{Individual Models} 
&TransE & 0.356  & 0.256 & 0.419 & 0.531 & 0.336 & 0.243 & 0.369 & 0.524 & 0.300 & 0.212 & 0.340 & 0.469 \\
&DistMult & 0.443 & 0.412 & 0.453 & 0.504 &  0.343 & 0.249 & 0.380 & 0.533 & 0.322 & 0.238 & 0.359  & 0.486 \\
&RotatE & 0.387  & 0.376 & 0.392 & 0.409 & 0.266 & 0.188 & 0.289 & 0.422 & 0.322 & 0.238 & 0.359 & 0.493 \\
&ComplEx & 0.487  & 0.443 & 0.503  & 0.573 & 0.265 & 0.186 & 0.290 & 0.422 & 0.323 & 0.237 & 0.362 & 0.492  \\
&AttE & \underline{0.491}  & \underline{0.444} & \underline{0.507} & \underline{0.583} & \underline{0.359} & \underline{\textbf{0.264}} & \underline{0.395} & \underline{0.548} & \underline{0.377} & \underline{0.292} & \underline{0.419} & \underline{0.539}\\
&AttH & 0.482 & 0.434 & 0.502 & 0.576 & 0.356 & 0.262 & 0.393 & 0.546 & 0.366 & 0.279 & 0.412 & 0.532 \\
\hline
\hline
\multirow{2}{*}{Our models} 
& SEPA & 0.480 & 0.436 & 0.498 & 0.570 &  0.354 & 0.259 & 0.390 & 0.545 & 0.347 & 0.260 & 0.387 & 0.520 \\
& SEA & \textbf{0.500} & \textbf{0.454} & \textbf{0.518} & \textbf{0.591} & \textbf{0.360} & \textbf{0.264} & \textbf{0.398} & \textbf{0.549} & \textbf{0.384} & \textbf{0.294} & \textbf{0.432} & \textbf{0.554}\\
\hline
\hline
\multirow{1}{*}{Ablation} 
&  SE & 0.495 & 0.448 & 0.513 & 0.587 & 0.353 & 0.259 & 0.389 & 0.542 & 0.381 & 0.292 & 0.427 & 0.548 \\
\hline
\end{tabular}
\end{adjustbox}
\caption{Link prediction evaluation on datasets for d=500. 
}
\label{tab:main_table_high}

\end{table*}

\subsection{ Further analyses}
We  additionally make a series of further analyses to evaluate the performance of our attention-base combination function. First, we want to show that our model is able to increase the precision of predictions for both symmetric and antisymmetric relations.
Table \ref{tbl:comp_rel} shows the H@1 results in WN18RR, in the low-dimensional setting of SEPA, compared to the individual combined KGE. Further results on H@10 can be found in Appendix \ref{h@10_appendix}. 
For example, if we look at the symmetric relation \textit{derivationally related form}, we can see that the H@1 of TransE is very low when compared to the one of ComplEx and DistMult, and yet, our model was able to improve this metric. 
Similarly, when we look at an antisymmetric relation (e.g., \textit{member of domain usage}) we have the opposite situation, having high performance for TransE and a lower one for ComplEx and DistMult. 
The intuition is that the attention base combination can effectively give more importance to the best models for the specific kind of relation involved in the query.
Such intuition is reinforced in Figure \ref{plot_rel}, which shows the (averaged) attention value among the individual models for the above-mentioned relations. 
It shows that the attention function can effectively select the correct proportion among the
models for the two different kinds of relations.

Besides, the importance of the attention function is highlighted by our ablation study, which consists of turning off the attention from our best models, SEPA at dimension 32, and SEA at dimension 500. 
We obtained two new versions of the models, namely \textbf{SEP} and \textbf{SE}. Tables \ref{tab:main_table_low}, \ref{tab:main_table_high} show that SEPA and SEA outperform SEP and SE.

\begin{table}[H]
\centering
\small
\caption{Comparison between our proposed models and the ensemble models proposed in MulDE \cite{wang2021mulde}. Values marked `-' were not reported in the original paper.}
\begin{tabular}{|l|ccc|ccc|}
\hline
\multirow{2}{*}{\textbf{Model}} & \multicolumn{3}{c|}{\textbf{WN18RR}}& \multicolumn{3}{c|}{\textbf{FB15k-237}}\\ \cline{2-7}
& MRR & H@1 & H@10 & MRR & H@1 & H@10\\ \hline
\multirow{1}{*}{MulDE \textsubscript{1}} 
 & 0.481 & \underline{0.433} & 0.574 & \underline{0.328} & \underline{0.237} & \underline{0.515}\\
\multirow{1}{*}{MulDE \textsubscript{2}} 
& \underline{0.482} & 0.430 & \textbf{0.579} & - & - & - \\
\multirow{1}{*}{SEPA \textsubscript{1}} 
& 0.481 & 0.441 & 0.562 & 0.332 & 0.243 & 0.509 \\
\multirow{1}{*}{SEPA \textsubscript{2}} 
& \textbf{0.491} & \textbf{0.448} & 0.572 & \textbf{0.344} & \textbf{0.252} & \textbf{0.528} \\
\hline
\end{tabular}
\label{tbl:comp_MuldE}
\end{table}

\subsection{Comparison With Ensemble Models}

We further compare our models to MulDE \cite{wang2021mulde}, which uses 64 to 512 dimensional embeddings for the teachers, and 8 to 64 dimensional ones for the junior embedding. We selected the best version of their model, MulDE\textsubscript{1} having 32 and 64 as junior and teachers dimensions, respectively, and, for a fair comparison, MulDE\textsubscript{2}, having dimension 64 for both junior and teachers embeddings.
We brought our models to their setting, testing SEPA at dimension 32 (SEPA\textsubscript{1}) and 64 (SEPA\textsubscript{2}) for both WN18RR and FB15k-237. 
Table \ref{tbl:comp_MuldE} shows that apart from the value of H@10 in WN18RR, our models substantially outperform such baselines, 
 having for MRR and H@1 up to 3,46\% relative improvements. 
Besides, we notice that when increasing the number of dimensions the performance of MulDE on H@1 in WN18RR decreases, and the ones of MRR and H@10 slightly increase. 
On the other hand, our models can substantially improve their performance when increasing the number of dimensions.

\subsection{Comparison With Models On Ultrahyperbolic Space}
Additionally, we compared our models against the best versions of UltraE \cite{xiong2022ultrahyperbolic}.
Even though we did not utilize Ultrahyperbolic space as a sophisticated manifold containing several sub-manifolds, our models get competitive results to the state-of-the-art in the Ultrahyperbolic space (Table \ref{tbl:comp_UltraE}).
In particular, SEPA gets competitive results in low-dimensions, while SEA in high-dimensions. 

One may consider using our idea to integrate approaches such as \cite{xiong2022ultrahyperbolic, gu2018learning} with other baselines.
However, due to their involved multiple geometric spaces, such integration will require a substantial revision of combinations of transformations and, hence, is left for future work.

\begin{table}[t]
\centering
\small
\caption{Comparison between our proposed models and the best UltraE \cite{xiong2022ultrahyperbolic} in WN18RR. Best score in bold and second best underlined.}
\begin{tabular}{|l|l|cccc|}
\hline
\multirow{2}{*}{\textbf{}} & \multirow{2}{*}{\textbf{Model}} & \multicolumn{4}{c|}{\textbf{WN18RR}} \\ \cline{3-6}
&  & MRR & H@1 & H@3 & H@10\\ \hline
\multirow{2}{*}{d=32} 
& UltraE \textsubscript{(q=4)} & \textbf{0.488} & \underline{0.440} & \textbf{0.503} & \underline{0.558}\\
& SEPA & \underline{0.481} & \textbf{0.441} & \underline{0.496} & \textbf{0.562}\\
& SEA & 0.466 & 0.425 & 0.482 & 0.542\\
\hline
\hline
\multirow{2}{*}{d=500} 
& UltraE \textsubscript{(q=40)} & \textbf{0.501} & \underline{0.450} & \underline{0.515} & \textbf{0.592}\\
& SEPA & 0.480 & 0.436 & 0.498 & 0.570\\
& SEA & \underline{0.500} & \textbf{0.454} & \textbf{0.518} & \underline{0.591} \\
\hline
\end{tabular}
\label{tbl:comp_UltraE}
\end{table}
\section{Conclusion}
In this paper, we propose a new approach that facilitates the combination of the query representations from a wide range of popular knowledge graph embedding models, designed in different spaces such as Euclidean, Hyperbolic, ComplEx, etc. We presented a spherical approach together with attention to queries to capture heterogeneous logical and structural patterns. We presented a theoretical analysis to justify such characteristics in expressing and inferring patterns and provided experimental analysis on various benchmark datasets with different rates of patterns to show our models uniformly perform well in link prediction tasks on various datasets with diverse characteristics in terms of patterns.
Our ablation studies, relation analysis on WN18RR and analysis of the learned attention values show our models mainly take the advantage of the best-performing models in link prediction tasks. By doing that, we achieved state-of-the-art results in Euclidean and Hyperbolic spaces.

In future work, we will combine various manifolds besides combining the queries in knowledge graph embedding. 
 Additionally, the proposed approach could be applied to other tasks. For example, it could be possible to use an attention mechanism to combine multi-hop queries computed using different complex query answering methods~\cite{ren2020beta, DBLP:conf/iclr/RenHL20}.
\begin{acks}
This work has received funding from the following projects:
The European Union’s Horizon 2020 research and innovation program under the Marie Skłodowska-Curie grant agreement No: 860801; 
BMWi Servicemeister (01MK20008F);
DFG - COFFEE (STA 572\_15-2); and
DFG Excellence Strategy in the Clusters of Excellence IntCDC at the University of Stuttgart, RP 20.
\end{acks}

%so that the information contained therein can be more easily collected
%during the article metadata extraction phase, and to ensure
%consistency in the spelling of the section heading.

% Authors should not prepare this section as a numbered or unnumbered {\verb|\section|}; please use the ``{\verb|acks|}'' environment.

\clearpage
\newpage
\bibliographystyle{ACM-Reference-Format}
\bibliography{references}

% \appendix

% \section{APPENDIX}
\appendix
\clearpage
\section{Appendix}
This appendix includes the proof of the propositions proposed in the paper, followed by optimal hyperparameter settings, additional related works, further explanation of Table 1, and more experimental analysis.

\subsection{Proofs} \label{proof_appendix}

\paragraph{\textbf{Proof of Proposition \ref{proposition:spherical:answerbetween}}}

This proposition states that a vector $\boldsymbol{a}_1$ is a solution of the minimization problem if and only if $\boldsymbol{a}_1$ lies between of the vector query embeddings $\boldsymbol{q}_1$ and $\boldsymbol{q}_2$ of a query $q$.
Without loss of generality, we assume that $\boldsymbol{q}_1$ and $\boldsymbol{q}_2$ lie in the x-axis.
The equivalence is proved in both directions.

\begin{enumerate}
\item If $\boldsymbol{a}_1$ is the solution of the minimization problem, we need to prove that $\boldsymbol{a}_1$ lies between $\boldsymbol{q}_1$ and $\boldsymbol{q}_2$. Assume that $\boldsymbol{a}_1$ does not lie between $\boldsymbol{q}_1$ and $\boldsymbol{q}_2$. By the triangle inequality, $p(\boldsymbol{q}_1, \boldsymbol{a}_1) + p(\boldsymbol{q}_2, \boldsymbol{a}_1) \geq p(\boldsymbol{q}_1, \boldsymbol{q}_2)$. Since $p(\boldsymbol{q}_1, \boldsymbol{q}_E) + p(\boldsymbol{q}_2, \boldsymbol{q}_E) = p(\boldsymbol{q}_1, \boldsymbol{q}_2)$, it follows that $\boldsymbol{a}_1$ is not the minimum. This contraction comes from assuming that $\boldsymbol{a}_1$ does not lie between $\boldsymbol{q}_1$ and $\boldsymbol{q}_2$. Hence, $\boldsymbol{a}_1$ lies between $\boldsymbol{q}_1$ and $\boldsymbol{q}_2$.

\item Let $\boldsymbol{a}_1$ be in between of $\boldsymbol{q}_1$ and $\boldsymbol{q}_2$, and $\boldsymbol{a}_2$ be an arbitrary vector that does not lie between $\boldsymbol{q}_1$ and $\boldsymbol{q}_2$. By the triangle inequality, $p(\boldsymbol{q}_1, \boldsymbol{a}_2) + p(\boldsymbol{q}_2, \boldsymbol{a}_2) \geq p(\boldsymbol{q}_1, \boldsymbol{q}_2)$. Since $\boldsymbol{a}_1$ lies between $\boldsymbol{q}_1$ and $\boldsymbol{q}_2$, it follows that $p(\boldsymbol{q}_1, \boldsymbol{q}_2) = p(\boldsymbol{q}_1, \boldsymbol{a}_1) + p(\boldsymbol{q}_1, \boldsymbol{a}_1)$. Substituting $p(\boldsymbol{q}_1, \boldsymbol{q}_2)$ with the right side of the last equation in the last inequality, we conclude that $p(\boldsymbol{q}_1, \boldsymbol{a}_2) + p(\boldsymbol{q}_1, \boldsymbol{a}_2) \geq p(\boldsymbol{q}_1, \boldsymbol{a}_1) + p(\boldsymbol{q}_2, \boldsymbol{a}_1)$. Hence, $\boldsymbol{a}_1$ is a solution of the minimization problem. \qed
\end{enumerate}

\paragraph{\textbf{Proof of Proposition \ref{subsumption}}}
Here, we prove that our hyperbolic version of our model SEPA subsumes AttH. 
We start with the sketch of the proof on the example of AttH, and then present the complete proof in detailed steps in a general case.
Because every transformation used in AttH is also used in SEPA with attention values, if we set the attention values corresponding to the transformations used in AttH, and set the other attention values to zero, then SEPA represents AttH scoring. Therefore, AttH is a special case of SEPA. 

Similar points exist for ComplEx, TransE and DistMult. 

Here we prove mathematically that SEPA can represent the same query to AttH.
let us assume that we combine 
the queries $\boldsymbol{q_{AttH}}$,  $\boldsymbol{q_{TransE}}$, $\boldsymbol{q_{DistMult}}$, 
and $\boldsymbol{q_{ComplEx}}$, 
which are the query representations of AttH, TransE, DistMult, and ComplEx respectively.
According to \autoref{eq:attentionquery}, we have
$\boldsymbol{q_{SEPA}} = \boldsymbol{a}_1\, \boldsymbol{q_{AttH}} + \,\boldsymbol{a}_2\, \boldsymbol{q_{TransE}} + \,\boldsymbol{a}_3\, \boldsymbol{q_{Distmult}} + \,a_4\, \boldsymbol{q_{ComplEx}}$.
By setting relation embeddings of ComplEx and DistMult to zero, we have
$\boldsymbol{q_{DistMult}} = \boldsymbol{q_{ComplEx}} = \boldsymbol{0}$.
Now, we need to find a way to cancel the query of TransE by setting the attention value of the TransE query to zero.
$\boldsymbol{a}_2$ will be close to zero if $\boldsymbol{r_{TransE}} = -M \boldsymbol{w}$ (in \autoref{eq:attentioncompute}) ($M>0$ is a sufficiently large number). 
Therefore, the query of TransE will be canceled in $\boldsymbol{q_{SEPA}}$, and we then have 
$\boldsymbol{q_{SEPA}} \approx \boldsymbol{a}_1 \boldsymbol{q_{AttH}}$.
Note that $\boldsymbol{a}_1$ is close to one if $\boldsymbol{w}$ (in \autoref{eq:attentioncompute}) is a large vector almost parallel to $\boldsymbol{q_{AttH}}$. Therefore, we have $\boldsymbol{q_{SEPA}} \approx \boldsymbol{q_{AttH}}$. 
Because the query is approximated, the score is also approximated.
Similarly, we can prove that SEPA can approximate the query of TransE, ComplEx, and DistMult, thus their score as well.

A similar process is also applicable for proof of subsumption between SEA and AttE, where AttE is a special case of SEA. Therefore, SEA can represent any scoring presented by AttE, and this completes the proof.

\paragraph{\textbf{Proof of Corollary \ref{inferpatterns}}}
This is a direct conclusion of Proposition \ref{subsumption}, thus it is a corollary. In Proposition \ref{subsumption}, we prove that SEPA and SEA subsume TransE, DistMult, RotatE, ComplEx and AttH (AttE), and also we know from \cite{nayyeri20205} that if a model A subsumes a model B, then the model A can infer all patterns that the model B can infer. Therefore, SEPA and SEA can infer the patterns that TransE, DistMult, RotatE, ComplEx, and AttH (AttE) can infer. 
It has been already proven in \cite{sun2019rotate} that the models are capable of inferring anti-symmetry, symmetry, composition, and inversion.

\begin{table}[ht!]
\centering
\caption{Comparison of H@10 for WN18RR relations. TE = TransE, CE = ComplEx, DM = DisMult}
\begin{tabular}{l|llll}
\hline
\multirow{1}{*}{Relation} & \multirow{1}{*}{TE}& \multirow{1}{*}{CE}& \multirow{1}{*}{DM} & \multirow{1}{*}{SEPA}
\\\hline
member meronym & \textbf{\underline{0.427}} & 0.223 & 0.130 & 0.409\\
hypernym & 0.197 & 0.124 & 0.065 & \textbf{0.277}\\
has part & \textbf{\underline{0.334}} & 0.227 & 0.142 & 0.323\\
instance hypernym & 0.480 & 0.426 & 0.221 & \textbf{0.500}\\
member of domain region & 0.417 & 0.229 & 0.154 & \textbf{0.481}\\
member of domain usage & 0.423 & 0.231 & 0.104 & \textbf{0.458}\\
synset domain topic of & 0.447 & 0.241 & 0.162 & \textbf{0.461}\\
also see & \textbf{\underline{0.723}}& 0.625 & 0.616 & 0.714\\
derivationally related form & 0.958 & 0.957 & 0.946 & \textbf{0.966}\\
similar to & \textbf{\underline{1.000}} & \textbf{\underline{1.000}} & \textbf{\underline{1.000}} & \textbf{1.000}\\
verb group & 0.962 & \textbf{\underline{0.974}} & \textbf{\underline{0.974}} & \textbf{0.974}\\
\hline

\end{tabular}
\label{tbl:comp_rel1}
\end{table}

\subsection{Best hyperparameters per model and per dataset}  \label{hyperparam_appendix}
\autoref{tab:hyperparamW}, \ref{tab:hyperparamF}, \ref{tab:hyperparamN} specifies the hyperparameter lists corresponding to the WN18RR, FB15k-237 and NELL-995-h100 respectively.

\begin{table}[ht!]
\centering
\small
\resizebox{\columnwidth}{!}{
\begin{tabular}{|l|cccccccc|}
\hline \multirow{1}{*}{\textbf{Model}} & \multicolumn{8}{c|}{\textbf{WN18RR}} \\ \cline{2-9}
& m & lr & o & n & b & dt & ar & dn \\ \hline
\multirow{1}{*}{SEPA (d=32)}& TCD & 0.001 & Adam & 250 & 500 & single & yes & yes\\
\multirow{1}{*}{SEPA (d=500)}& TCD & 0.001 & Adam & 250 & 500 & single & yes & yes\\
\multirow{1}{*}{SEA(d=32)}& TCD  &0.001 & Adam & 250 & 100 & single & yes & yes\\
\multirow{1}{*}{SEA(d=500)}& ALL  &0.001 & Adam & 250 & 100 & single & no & no\\
\hline
\end{tabular}

}
\caption{Best Hyperparameters for WN18RR. T = TransE, D = DistMult, C = ComplEx, R = RotatE, AttH(Reflection) = A. ALL = all models combined}
\label{tab:hyperparamW}

\resizebox{\columnwidth}{!}{
\begin{tabular}{|l|cccccccc|}
\hline \multirow{1}{*}{\textbf{Model}} & \multicolumn{8}{c|}{\textbf{FB15k-237}} \\ \cline{2-9}
& m & lr & o & n & b & dt & ar & dn \\ \hline
\multirow{1}{*}{SEPA (d=32)}& TCD & 0.05 & Adagrad & 250 & 500 & double & yes &no\\
\multirow{1}{*}{SEPA (d=500)}& ALL & 0.05 & Adagrad & 250 & 100 & double & no & no  \\
\multirow{1}{*}{SEA(d=32)}& ALL & 0.1 & Adagrad & 250 & 500 & single & no & no \\
\multirow{1}{*}{SEA(d=500)}& ALL & 0.1 & Adagrad & 250 & 500 & single & no & no \\
\hline
\end{tabular}
}
\caption{Best Hyperparameters for FB15k-237.}
\label{tab:hyperparamF}

\resizebox{\columnwidth}{!}{
\begin{tabular}{|l|cccccccc|}
\hline \multirow{1}{*}{\textbf{Model}} & \multicolumn{8}{c|}{\textbf{NELL-995-h100}} \\ \cline{2-9}
& m & lr & o & n & b & dt & ar & dn \\ \hline
\multirow{1}{*}{SEPA (d=32)}& ALL & 0.001 & Adam & 250 & 100 & single & no &no\\
\multirow{1}{*}{SEPA (d=500)}& ALL & 0.001 & Adam & 250 & 500 & single & no &no \\
\multirow{1}{*}{SEA(d=32)}& ALL & 0.001 & Adam & 250 & 500 & single & no & no \\
\multirow{1}{*}{SEA(d=500)}& ALL & 0.001 & Adam & 250 & 500 & single & yes & no  \\
\hline
\end{tabular}
}
\caption{Best Hyperparameters for NELL-995-h100.}
\label{tab:hyperparamN}

\end{table}

\subsection{Related Work: combination between machine learning models \textit{including} KGE} \label{rel_work_app}
We review the related work corresponding to the general approaches in machine learning, including embedding that combines different models.

DuEL \cite{joshi2022ensemble} exploits embedding models for classifying facts to be either true or false, rather than ranking them. 
Starting from a tail query $(h,r,?)$, it uses an embedding model to obtain the top $k$ list of predicted answers and feeds different classifiers (e.g., LSTM, CNN) to label each answer as true or false. 
Finally, the predictions are ensembled using different techniques.
A similar approach \cite{qudushybridfc}, also proposed an ensemble-based framework for fact-checking. Starting from a triple $(h,r,t)$, it runs three different methods: (1) a text-based approach, (2) a KGE model, and (3) a path-based approach. 
It concatenates the outputs and lets a neural network compute a final veracity score.
On the other hand, in our work, we propose to combine query representations of different KGE models.

\subsection{Further Explanation of \autoref{tbl:baselines}} \label{explaintbl1}
Here, we present a further explanation of the results in \autoref{tbl:baselines}, and show how to compute the constraints.
\begin{itemize}
    \item \textit{TransE/Symmetric}: According to the \autoref{tbl:baselines}, TransE cannot model a symmetric pattern. 
    To show this, we take a triple $(h,r,t)$ and its symmetry $(t,r,h)$. To model both triples in the vector space by TransE, we need to have 
    \begin{equation*}
        \boldsymbol{h + r = t}, \,\,\,\boldsymbol{t + r = h}.
    \end{equation*}
    Combining the two equations leads to $\boldsymbol{r = 0}$.\. 
    This is counted as incapability for modeling symmetry by TransE because a null vector for relation embedding leads to the same embeddings for all entities connected by the relation.
This is in contradiction with the assumption of embedding models to assign unique vectors to each entity in the KG.
    \item \textit{TransE/AntiSymmetry}:
    Lets $r$ be antisymmetry.
    Thus, if $(h,r,t)$ is true,
    we have $\boldsymbol{h + r = t}$. This trivially leads to  $\boldsymbol{t + r \neq h}$ without using further constraint. 
    Note that we use the assumption that entity embeddings are unique in the vector space.
\item \textit{TransE/Hierarchy}: According to the \autoref{tbl:baselines}, TransE cannot model hierarchical patterns.
To show this, we take a small part of a tree as a hierarchical structure. 
We consider the root ($h$) and two children of the root node entity $t_1, t_2$.
The triples $(h,r,t_1)$ and  $(h,r,t_2)$ are valid.
To model both triples in the vector space by TransE, we need to have 
    \begin{equation*}
        \boldsymbol{h + r = t_1}, \,\,\,\boldsymbol{h + r = t_2}.
    \end{equation*}
Comparing the two equations, we conclude that $\boldsymbol{t_1 = t_2}.$
This is counted as incapability for modeling hierarchy by TransE because all entities embeddings are assumed to be unique in the vector space which is not the case for hierarchy.
\end{itemize}

A similar calculation is required for other patterns/models which we do not go through because the process is similar to TransE.

Given a pattern, 
a model with fewer constraints has better inference capability compared to the models with more constraints.
According to the \autoref{subsumption} and \ref{inferpatterns}, our model can take advantage of each model without being affected by their incapability due to using a relation-specific attention mechanism. 

\subsection{H@10 per relation} \label{h@10_appendix}
\autoref{tbl:comp_rel1} presents the performance of our model and different base models considering the metric Hits@10 on each relation of the WN18RR dataset. 
We have a similar observation to the results on Hits@1.
In most cases, our model outperforms the base models. 
For a few relations, our model gets slightly lower performance than the base models.
We hypothesize that might be related to numerical optimization, or our model aimed to increase the overall accuracy for all relations, so for the relations with less frequencies, our model gets slightly lower performance to have overall higher performance in all relations when increase in performance of one relation leads to decrease the performance of other relation. 

\subsection{Comparison with TuckER and MuRP}\label{comp_tuck}
Here, we present further analysis of our models with MuRP and TuckER. 
In particular, as shown in Table \ref{tbl:comp_Tucker} we compared MuRP with the low-dimensional versions of our models, and TuckER with the high-dimensional ones. 
The reason is that MuRP is a hyperbolic model, hence its best performances are shown in low-dimensional space, while TuckER is a Euclidean one and shows the best performances in a high-dimensional space.

We observe that SEPA outperforms MuRP in all metrics, having for example a relative improvement of around 5\% in the H@1 metric of the WN18RR dataset. 
Besides, SEA outperforms TuckER for all metrics of WN18RR and most metrics of FB15k-237. In particular, it obtains around 12\% relative improvements in the H@1 of WN18RR. 
On the other hand, it obtains around 0.75\% relative worsening in the H@1 of FB15k-237.

Overall, we observe that our models were able to outperform such baselines even though they were not included in the combination.

\begin{table}[t]
\centering
\small
\caption{Comparison with MuRP, against models of dimension 32, and TuckER, against models of dimension 500, which results were taken from \cite{chami2020low}. Best score in bold and second best underlined. }
\setlength\tabcolsep{3.4 pt}
\begin{tabular}{|l|cccc|cccc|}
\hline
\multirow{2}{*}{\textbf{Model}} & \multicolumn{4}{c|}{\textbf{WN18RR}} & \multicolumn{4}{c|}{\textbf{FB237}} \\ \cline{2-9}
& MRR & H@1 & H@3 & H@10 & MRR & H@1 & H@3 & H@10\\ \hline
\multirow{1}{*}{MuRP} 
& 0.465 & 0.420 & 0.484 & \underline{0.544} & 0.323 & 0.235 & 0.353 & 0.501\\
\multirow{1}{*}{SEPA} 
& \textbf{0.481} & \textbf{0.441} & \textbf{0.496} & \textbf{0.562} &  \textbf{0.332} & \textbf{0.243} & \textbf{0.363} & \textbf{0.509} \\
\multirow{1}{*}{SEA} 
& \underline{0.468} & \underline{0.430} & \underline{0.485} & 0.538 & \underline{0.326} & \underline{0.238} & \underline{0.356} & \underline{0.504}\\
\hline
\hline
\multirow{1}{*}{TuckER} 
& 0.470 & \underline{0.443} & 0.482 & 0.526 & \underline{0.358} & \textbf{0.266} & \underline{0.394} & 0.544\\
\multirow{1}{*}{SEA} 
& \underline{0.480} & 0.436 & \underline{0.498} & \underline{0.570} &  0.354 & 0.259 & 0.390 & \underline{0.545}\\
\multirow{1}{*}{SEA} 
& \textbf{0.500} & \textbf{0.454} & \textbf{0.518} & \textbf{0.591} & \textbf{0.360} & \underline{0.264} & \textbf{0.398} & \textbf{0.549}\\
\hline
\end{tabular}
\label{tbl:comp_Tucker}
\end{table}

\end{document}